\RequirePackage[2020-02-02]{latexrelease}

\documentclass{clv3}
\usepackage[hyphens]{url}
\usepackage{xcolor}
\definecolor{darkblue}{rgb}{0, 0, 0.5}
\usepackage{algorithmic}
\usepackage{epigraph,lscape}
\usepackage{latexsym,amsmath,amssymb}
\usepackage{textcomp}
\usepackage{ragged2e}  
\usepackage[caption=false,font=normalsize,labelfont=sf,textfont=sf]{subfig}
\let\citet=\cite
\usepackage{verbatim}
\usepackage{soul}
\usepackage{versions}\excludeversion{ignore}
\usepackage{comment,todonotes}
\usepackage{colortbl}
\usepackage{threeparttable}
\usepackage{multirow}
\usepackage[figuresright]{rotating}
\issue{1}{1}{2023}
\runningtitle{Domain-Independent Deception}
\runningauthor{Verma, Dershowitz, Zeng, Boumber \& Liu}
\pageonefooter{Action editor: }
\title{Domain-Independent Deception:\\ A New Taxonomy and Linguistic Analysis%
\thanks{This is a thoroughly revised version of a 2022 arXiv draft containing substantial new material.}}
\author{Rakesh Verma} 
\affil{Department of Computer Science \\University of Houston}
\author{Nachum Dershowitz}
\affil{School of Computer Science\\ Tel Aviv University}
\author{Victor Zeng}
\affil{University of Houston and InstaBase}
\author{Dainis Boumber}
\affil{University of Houston}
\author{Xuting Liu\thanks{Work performed at the University of Houston and U. C. Berkeley.}}
\affil{University of California -- Berkeley}
\begin{document}
\maketitle

\begin{abstract}
    Internet-based economies and societies are drowning in deceptive attacks. These attacks take many forms, such as fake news, phishing, and job scams, which we call ``domains of deception.'' Machine-learning and natural-language-processing  researchers have been attempting to ameliorate this precarious situation by designing domain-specific detectors. Only a few recent works have considered domain-independent deception. We collect these disparate threads of research and investigate domain-independent deception. First, we provide a new computational definition of deception and break down deception into a new taxonomy. Then, we analyze the debate on linguistic cues for deception and supply guidelines for systematic reviews. Finally, we investigate common linguistic features and give evidence for knowledge transfer across different forms of deception. 

\medskip  
\noindent{\bf Keywords:} \rm   Automatic/computational deception detection, 
cross domain, domain independent,  
email/message scams,
fake news, 
meta-analysis,  
opinion spam, 
phishing,
social engineering attacks,
systematic review,
text analysis.
\end{abstract}

\section{Introduction} \label{sec:introduction}
History is replete with famous lies and deceptions. Examples include: P. T. Barnum, Nicolo Machiavelli, Sun Tzu, Operation Mincemeat, and the Trojan Horse~\cite{encyclopedia_deception}. A chronology of deception is included in~\cite{encyclopedia_deception}. More recently, 
the proliferation of deceptive attacks such as fake news, phishing, and disinformation is rapidly eroding trust in Internet-dependent societies. The situation has deteriorated so much that 45\% of the US population believes the 2020 US election was stolen.\footnote{\url{https://www.surveymonkey.com/curiosity/axios-january-6-revisited}.}
 
Social-media platforms have come under severe scrutiny regarding how they police content. Facebook and Google are partnering with independent fact-checking organizations that typically employ manual fact-checkers. 

Natural-language processing (NLP) and machine learning (ML) researchers have joined the fight by designing fake news, phishing, and other kinds of domain-specific detectors. 

Building single-domain detectors may be sub-optimal. Composing them sequentially requires more time, and composing them in parallel requires more hardware. Moreover, building single-domain detectors means one can only react to new forms of deception after they emerge. 

Our goal here is to spur research on \emph{domain-independent} deception. Unfortunately, research in this area is currently hampered by the lack of computational definitions and taxonomy, high-quality datasets, and  systematic approaches to domain-independent deception detection. Thus, results are neither generalizable nor reliable, leading to much confusion. 

Accordingly, we make the following contributions:
\begin{itemize}
\item We propose a new computational definition and a new comprehensive taxonomy of deception. 
    (We use the unqualified term ``deception'' for the domain-independent case. When the goals of the deception  are unclear,  we refer to ``lies.'')
\item We examine the debate on linguistic deception detection, identify works that demonstrate the challenges that must be overcome to develop domain-independent deception detectors and examine them critically. 
\item  We conduct linguistic analysis of several detection datasets for general cues and find several statistically significant ones. 
\item We conduct deep learning experiments of deception sets and study correlations in performance for pairs of datasets. 
\end{itemize}

We hope that this article, besides scrutinizing the claims on general linguistic signals for deception, will aid those planning to conduct systematic reviews. Google Scholar searches with phrase queries of the two forms: (a) ``guidelines for systematic literature reviews in X'' and (b) ``systematic review guidelines for X,'' where $X \in \{\textrm{machine learning, ML, natural language processing, NLP, nlp}\}$ 
returned nothing. 

This article is organized as follows: Section 2 presents a new definition of deception. Section~\ref{sec-taxo} introduces our new taxonomy. Section~\ref{sec-related} summarizes related work. Section~\ref{sec-guide} presents  guidelines for systematic reviews and Sections~\ref{sec-debate},~\ref{sec-args} the linguistic cues debate. Sections~\ref{sec-cues} and~\ref{sec-dl} describe our experiments, results, and analysis of domain-independent markers for deception. Finally, Section~\ref{sec-concl} presents some conclusions and directions for the future. 
The appendices provide the list of features tested and some preliminary significance testing of cues on four public deception datasets. 

\section{Definition} \label{sec-defn}

We first examine a general definition of deception, taken from~\cite{galasinski2000language}, intended to capture a wide variety of deceptive situations and attacks. 
\begin{definition}[Preliminary]
\textbf{Deception} is an intentional act of manipulation to gain compliance. Thus, it has at least one source, one target, and one goal. The source is intentionally manipulating the target into beliefs, or actions, or both, intended to achieve the goals. 
\end{definition}
\noindent

Since we are interested in automatic verifiability, we would like to modify this definition of deception and propose one that is computationally feasible. 
Because intentions are notoriously hard to establish, we will use the effect of exposing the manipulation/goals instead.  

Our revised definition is the following:
\begin{definition}[Deception]
\textbf{Deception} is an act of manipulation designed to gain compliance such that, exposing the manipulation or the goal(s) of compliance significantly decreases the chance of compliance. 
Thus, it has at least one source, one target, and one goal. 
The source is manipulating the target into  beliefs, or action, or both, intended to achieve the goals. 
\end{definition}

One might argue that the goals of deception should be harmful to an individual or organization. However, this would necessitate either a computational definition of harm or a comprehensive list of potential harms, which could be checked computationally and is, therefore, a less desirable alternative.


To formalize our definition, we borrow from the language of Markov decision processes. Let $A$ be an action taken by an actor, and let $C$ be a desired compliance state. 
We use $K(A,T)$ to denote the action $A$ plus the full and truthful explanation of the actor's \emph{relevant} private information to target $T$. We formalize (computational) deception using conditional probabilities as follows:
\begin{definition}[Computational Deception -- Formalized] \label{def:formal}
An action $A$ deceives target $T$ if $P(C~|~K(A,T)) < P(C~|~A)$.
\end{definition}

Moreover, we can quantify the degree to which $A$ is deceptive by the amount $\theta$, where $0 \leq \theta \leq 1$.
\begin{definition}[Computational Deception -- Quantified]
An action $A$ $\theta$-deceives $T$ if $P(C~|~K(A,T)) \leq P(C~|~A) - \theta$.
\end{definition}

In practice, practitioners can apply this by exposing the manipulation and/or goals and measuring the change in compliance rates. For example, a Florida woman recently sued Kraft alleging that the “ready in 3½ minutes” on the label of their microwavable Velveeta Shells \& Cheese is deceptive. To determine whether the claim is in fact deceptive, a researcher could present the product by itself to one group of random consumers and the product with an explanation that the 3½ minutes does not include the time to add water to another group. If there is a statistically significant decrease in purchases (which is the desired compliance) for the group with the explanation, then the claim is deceptive.

There is some work on finding out how good humans are at detecting certain kinds of deceptive attacks. For the detection capabilities of automatic detectors on specific domains of deception, one can look at surveys on fake news detection~\cite{sharma2019combating,zhou2019fake,zhou2020fake} and phishing detection~\cite{dasBA20}. 


\begin{figure*}[tp]
\centering
\includegraphics[width=1.0\textwidth]{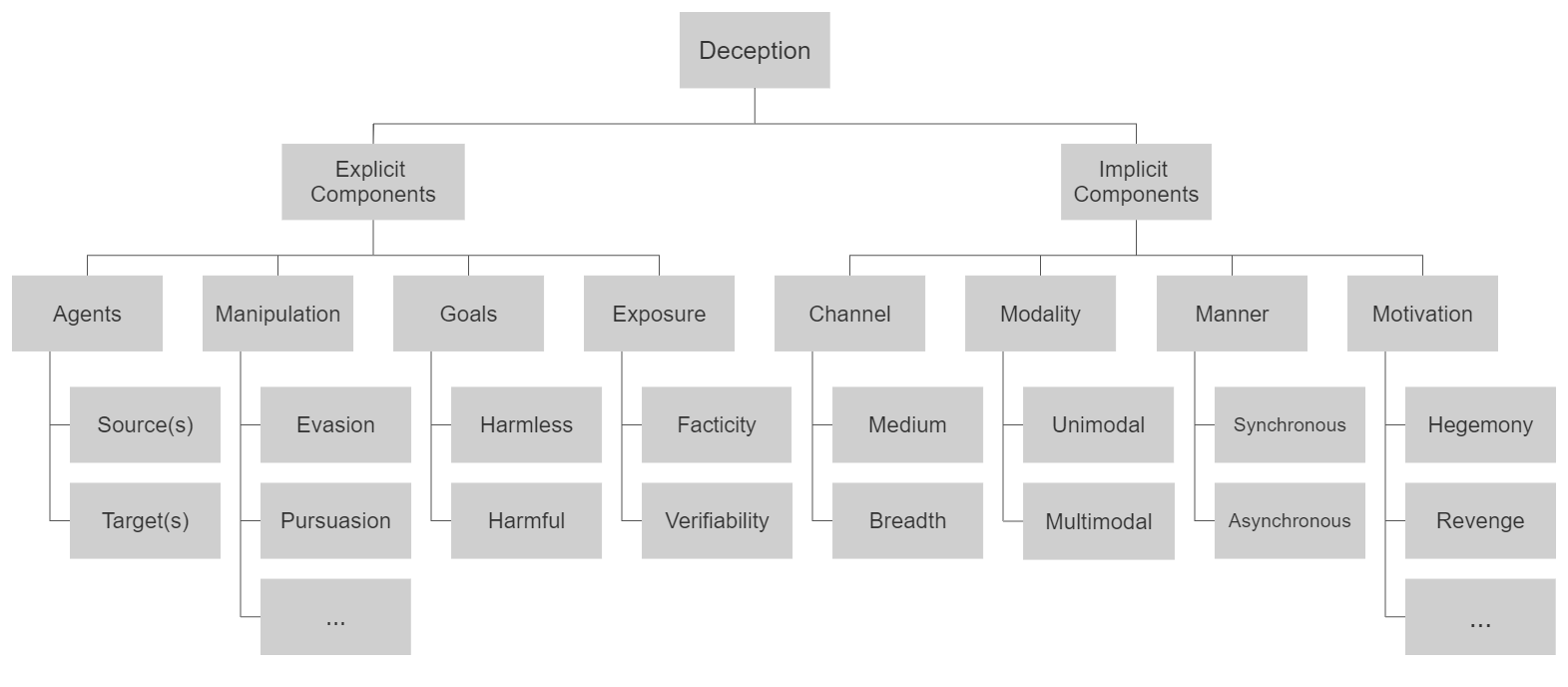}
\caption{The proposed deception taxonomy -- the full manipulation (or stratagem) and motivation subtrees are not shown.} \label{fig:taxonomy}
\end{figure*}


\section{Taxonomy and Examples}\label{sec-taxo}

There have been a few attempts at constructing taxonomies for fake news, phishing, or other forms of deception. 

Molina et al.~\citeyearpar{molina2019fake} give a taxonomy of {\em fake news}  with four dimensions: message and linguistic, sources and intentions, structural, and network. Kapantai et al.~\citeyearpar{kapantai2021systematic} conducted a systematic search for papers proposing taxonomies for disinformation and synthesized a taxonomy with three dimensions: factuality, motivation, and verifiability. 

No one, to our knowledge, has given a comprehensive taxonomy of real-world deception.

\subsection{The New Taxonomy}
We put forward a multi-dimensional taxonomy.
%
Under our definition, deception explicitly involves four elements: (1)  agents: the sources, and the targets, (2)  stratagems for manipulation, (3)  goals, and (4)  threat/mechanisms of exposure. 
These explicit elements can be further broken down as follows: 
\begin{enumerate}
\item[1)] \textit{Agents}. 
Rowe~\citeyearpar{rowe2006taxonomy} calls this category ``participant,'' and he further elaborates this into: 
(a) agent, who initiates the action, 
(b) beneficiary, who benefits,  
(c) object, what the action is done to, and 
(d) recipient, who receives the action. Rowe also includes experiencer (``who senses the action'') and instrument (``what helps accomplish the action'') components in this category, but we include them in the Channel category below. 
\item[1a)] {Sources}. This includes human (individual or group), bot, etc.,  or mixed, in other words, combinations such as a human assisted by a bot. The Sources category includes initiators and beneficiaries. 
\item[1b)] {Targets}. This includes humans (individual or group), automatic detectors, or both. For example, spam targets automatic detectors, and phishing targets humans, but needs to fool automatic detectors also. The Targets category includes the objects and the recipients. 
    \item[2)] \textit{Stratagems}. The stratagem subtree in the taxonomy includes two sub-taxonomies for persuasion and action, which we discuss below. We believe that persuasion is fundamental to deception since its goal is to change the reasoning of the target(s), with the deception's end goal of compliance. The action taxonomy is adapted from~\cite{rowe2006taxonomy}. It includes space, time, causality, quality, essence, and speech-act theory, which specifies the external and internal preconditions for the action. The persuasion taxonomy combines~\cite{cialdini2006} and~\cite{semeval23}. 
    \item[3)] \textit{Goals}.
        \item[3a)] {Harmless}: satire, parody, satisfying participation, as in a laboratory experiment where participants may be asked to lie, etc.
        \item[3b)] {Harmful}. This includes a wide range of objectives, such as stealing money or identity information, malware installation, manipulation of votes, planting fear, sowing confusion, initiating chaos, gaining an unfair edge in a competition ({e.g.}, swaying opinions and preferences on products), persuading people to take harmful actions, winning competitions/games, etc. We avoid the terms defensive and offensive since they are dependent on the perspective of the participants/agents. 
    \item[4)] \textit{Exposure}.
    \item[4a)] {Facticity}. Can we establish whether it is factual or not? For example, currently, we are unable to establish the truth or falsity of utterances such as, ``There are multiple universes in existence right now.'' 
    \item [4b)]{Verifiability}. Assuming facticity, how easy or difficult it is to verify whether it is legitimate or deceptive? Here, we are interested in machine or automatic verification. If a simple machine-learning algorithm can detect it with high recall and precision, we will deem it easy. 
\end{enumerate}

In addition, there are four implicit concepts in the definition: (1)  motivations behind the goals, (2)  communication channels or media, (3)  modality of deception, and (4)  manner or timeliness of the exchange. 
\begin{enumerate}
\item[1)] \textit{Motivation}. This is the rationale for the goals. The agents involved and their characteristics reveal the underlying motivations, which could be political hegemony (nation-states), religious domination, revenge (disgruntled employee), ideological gains, money, control, power, etc. 
    \item[2)] \textit{Channel}. This dimension includes two aspects:
        \item[2a)] Breadth: Whether the targets are a few specific individuals or detector types or broad classes of people/categories of detectors.
        \item[2b)] Media. How the deceptive capsule is conveyed to the target. Media also includes the experiencer and instrument components of Rowe~\citeyearpar{rowe2006taxonomy}.

    \item[3)] \textit{Modality}. This dimension refers to the presentation of deceptive content. It includes:

        \item[3a)] Unimodal. This includes only one type of modality such as (a) Gestural: body language is used to deceive, (b) Audio  (a.k.a.\@ verbal), (c) Textual ({e.g.}, SMS/email), and (d) Visual  ({e.g.}, images or videos).
        \item[3b)] Multimodal: combinations of different modalities.\\
     For example, audio-visual has both speech and visual components but lacks face-to-face communication in which gestures could facilitate deception.
    \item[4)] \textit{Manner/Timeliness}. 

        \item[4a)] Interactive/Synchronous. A real-time interview or debate is an interactive scenario. 
        \item[4b)] Non-interactive/Asynchronous.  An Amazon Mechanical Turker typing a deceptive opinion or essay is a non-interactive one. An asynchronous interaction can have multiple stages or steps some (but not all) of which may be synchronous. 
    \end{enumerate}


\subsubsection{Stratagems}\label{sec-strata}
Rowe's approach~\cite{rowe2006taxonomy} is based on linguistics. He states,  ``Each action has associated concepts that help particularize it, and these are conveyed in
language by modifiers, prepositional phrases, participial phrases, relative clauses, infinitives, and other constructs.'' These 
associated concepts are called `semantic cases'~\cite{fillmore68} in analogy to the syntactic cases that occur in some languages for nouns.
Rowe claims that ``every deception action can be categorized by an associated semantic case or set of cases.'' However, there is no canonical list of semantic cases in linguistics. Rowe prefers the detailed list from (Copeck et al, 1992), which he supplements with two important relationships from artificial
intelligence, the upward type-supertype and upward part-whole links, and two speech-act conditions from (Austin, 1975), to get 32 cases altogether. However, since we include his participant category in the Agents and Channel categories, we have only 26 subcategories in the Stratagems category. 
\begin{enumerate}
\item Space, which consists of: (a) direction, of the action, (b) location-at, where something occurred, (c) location-from, where something started, (d) location-to, where something finished,
(e) location-through, where some action passed through, and 
(f) orientation, in some space. 
\item Time, which is subdivided into: 
(a) frequency, of occurrence of repeated action, 
(b) time-at, time at which something occurred,
(c) time-from, the time at which something started,
(d) time-to, the time at which something ended, and 
(e) time-through, the time through which something occurred. 
\item Causality, which consists of: 
(a) cause, 
(b) contradiction, what this action opposes if anything, 
(c) effect, and 
(d) purpose. 
\item Quality, which is sub-divided into: 
(a)  accompaniment, an additional object associated with the action, 
(b) content, what is contained by the action object, 
(c) manner, the way in which the action is done, 
(d) material, the atomic units out of which the action is composed, 
(e) measure, the measurement associated with the action, 
(f) order, with respect to other actions, and 
(g) value, the data transmitted by the action (the software sense of the term). 
\item Essence, which consists of: 
(a) supertype, a generalization of the action type, and 
(b) whole, of which the action is a part. 
\item Speech-act theory, which is sub-divided into: 
(a) an external precondition on the action, and 
(b) an internal precondition, on the ability of the agent to perform the action. 
\end{enumerate}

\subsubsection{Persuasion}
We summarize the persuasion taxonomy in Table~\ref{tab:persuasion}. For this taxonomy, we adapt the SemEval 2023 Persuasion Task's categories~\cite{semeval23}, and Cialdini's~\citeyearpar{cialdini2006} persuasion principles, which are essentially persuasion techniques or strategies. The persuasion strategies taxonomy of~\cite{guerini2007taxonomy} is orthogonal to this taxonomy since their definition of persuasion is broader than ours, but we do include their specific strategies under Techniques. 
\begin{table}[tp]
    \centering 
    \caption{The Persuasion Taxonomy, adapted from~\cite{semeval23}, is a sub-taxonomy in the Deception Taxonomy.}
    \label{tab:persuasion}
    \begin{tabular}{|c|c|}
        \hline
      \bf  Category & \bf Description   \\\hline
        Justification & An argument made of two parts: a statement and a justification  \\ \hline
        Simplification & A statement is made that excessively simplifies a problem, usually \\
        & regarding the cause, the consequence or the existence of choices \\ \hline
        Distraction & A statement is made that changes the focus away \\
        & from the main topic or argument \\ \hline
   Call & The text is not an argument but an encouragement \\
   & to act or think in a particular way  \\ \hline
   Manipulative & Specific language/imagery is used or a statement is made \\
  Wording/Images & that is not an argument, and which contains words/phrases \\
   & that are either non-neutral, confusing, exaggerating, etc., \\
  &  to impact the reader, for instance emotionally \\ \hline
  Attack on & An argument whose object is not the topic of the conversation, \\
  Reputation &  but the personality of a participant, his experience and deeds, \\
  & typically to question and/or undermine his credibility \\ \hline
    \end{tabular}
\end{table}
The techniques used for each category are as follows (30 in total): 
\begin{itemize}
    \item Justification: Appeal to popularity, Appeal to authority/expert, Appeal to values (or Commitment~\cite{cialdini2006}), Appeal to fear/prejudice, Reciprocity~\cite{cialdini2006} (or Goal Balance~\cite{guerini2007taxonomy}), Scarcity~\cite{cialdini2006}, Reward, Appeal to relevant empirical evidence, Relevant Statistics, and Relevant Examples.
    \item Simplification: Causal oversimplification, False dilemma or no choice, and Consequential oversimplification.
    \item Distraction: Straw man, Red herring (includes irrelevant empirical evidence, statistics or examples), Whataboutism, Flag Waving, and Liking~\cite{cialdini2006}.
    \item Call:  Slogans, Social Proof~\cite{cialdini2006}, Appeal to time, and Conversation killer.
    \item Manipulative Wording/Images: Loaded language/images, Repetition, Exaggeration or minimization, and Obfuscation -- vagueness or confusion.
    \item Attack on reputation: Name calling or labeling, Doubt, Guilt by association, Appeal to hypocrisy, Questioning the reputation.
\end{itemize} 
   To the best of our knowledge, we are the first to use the following dimensions in a taxonomy of deception: target, persuasion, goal, dissemination, and timeliness. We add these to give a comprehensive view of deception, to aid in domain-independent deception detection, and to clarify and classify deception in all its different manifestations.
Such a comprehensive taxonomy will provide a solid foundation on which to build automatic and semi-automatic detection methods and training programs for the targets of deception. 

\subsection{Examples}
To demonstrate the applicability of this taxonomy, we give three examples. More discussion of stratagems and examples of cyber deception can be found in~\cite{rowe2006taxonomy}. 

Phishing is when attackers pretend to be from reputable companies to trick victims into revealing personal information. The agents are the attackers as initiators and the targets are the Internet/email users. The harmful goals include information or malware installation. Establishing the facticity is difficult if the attacker is determined.  The medium is the Internet/email. The breadth is high for phishing and narrower for spearphishing. The modality is text for phishing and audio for vishing. Images may also be used in phishing emails. The manner is non-interactive for phishing and interactive for vishing. Deliberate falsification and persuasion techniques such as authority, social proof, and reward or loss claims are employed in the stratagem. 

Fake news is manufactured and misleading information presented as news. Here the harmful goals include swaying opinion, sowing unrest, and division, etc. The sources could be individuals, organizations, or nation-states. The breadth could vary depending on how deep-pocketed and determined the source(s) is (are). The modality could be text, audio, images, or video. The manner is asynchronous. Fake news could employ a range of techniques in the action component of the stratagem: from deliberate falsification to evasion and the persuasion component could include authority, social proof, etc.

Fake reviews are reviews designed to give consumers a false impression of a product or business. The harmful goal is to convince consumers to buy their product or avoid a competitor's. The sources could be humans, bots, or their combinations. The targets are potential customers as well as the platform's fake review detector. The breadth is thus a broad range of people. While most fake reviews use only texts, deliberate attacks could be multi-modal, adding visuals and/or audio. Falsification and social proof are the main stratagems. Facticity and verifiability could vary depending on the stratagems used. The manner is asynchronous. 
\section{Related Work} \label{sec-related}
Deception has a vast social science literature. Hence, we focus on the most closely related work on computational deception, which can be categorized into: taxonomies, datasets, detection, and literature reviews. Of the latter, we focus here on reviews of linguistic deception detection. The DBLP query ``domain decepti'' on 30 August 2023 gave 30 matches of which 18 were deemed relevant. 

\subsubsection*{Remark} Unfortunately, previous researchers have generally left the term ``domain'' undefined. 
In~\cite{glenski2020towardsRDSM}, different social networks, such as Twitter and Reddit,  are referred to as domains. 
Hence, terms such as ``cross-domain deception'' in previous work could mean that the topics of essays or reviews are varied whereas the goals could stay pretty much the same.

\subsection{Taxonomies}
Whaley and Aykroyd \citeyearpar{whaley2007textbook} gave a taxonomy of perception in which deception was defined succinctly as ``other-induced misperception.'' The full definition given in~\cite{whaley2007textbook} is ``any attempt -- by words or actions -- intended to distort another person's or group's perception of reality.'' In~\cite{bell2017cheating} two groups were introduced as essential for deception: simulation (overt, showing the false) and dissimulation (covert, hiding what is real). They introduced three simulation techniques: mimicking, inventing, and decoying, and three dissimulation techniques: masking, repackaging, and dazzling. 

Dunnigan and Nofi~\citeyearpar{dunnigan2001victory} gave a taxonomy of deception in the military context. This included: concealment, camouflage, disinformation, lies, displays, ruses, demonstrations, feints, and insight. 

The most comprehensive previous taxonomy of deception, to our knowledge, is proposed in~\cite{rowe2006taxonomy}. It is inspired by linguistic case theory and includes 32 cases which are grouped into seven categories: space (six cases), time (five cases), participant (six cases), causality (four cases), quality (seven cases), essence (two cases), speech-act theory (two cases). Analyzing this taxonomy, we find that, except for the participant category, all the other categories fit neatly into the stratagems class for deception in our taxonomy. 

More recently, a few researchers have proposed more specialized taxonomies for what they call defensive deception~\cite{oluoha2021cutting,pawlickCZ19,pawlick2021game}. Some folksy and psychological taxonomies are given in~\cite{national1992mind}.

\subsection{Datasets}\label{sec-data-related-work}
Several datasets have been collected for studying lies. However, researchers have not carefully delineated the scope by considering the goals of the deception. There is also another potentially more serious issue: some datasets are constructed by asking participants to lie in a laboratory setting, where there are no consequences and no incentive to lie. We will refer to them as \textit{Lab Datasets}.  Others are constructed by collecting samples of real attacks. We call them \textit{Real-World Datasets}. Finally, there are some datasets in which data from laboratory settings are combined with real-world attack samples.  We call them \emph{Mixed Datasets}.

Lab Datasets include Zhou et al.~\citeyearpar{zhou}, wherein students were paired and one student in each pair was asked to deceive the other using messages. In~\cite{peres-rosas14}, researchers collected demographic data and 14 short essays (7 truthful and 7 false) on open-ended topics from 512 Amazon Mechanical Turkers (AMT). They tried to predict demographic information and facticity. We refer to this as the \textit{Open-Lies} dataset. In~\cite{perez2014cross}, researchers collected short essays on three topics: abortion, best friend, and the death penalty by people from four different cultural backgrounds. In~\cite{capuozzo}, truthful and deceptive opinions on five topics are collected in two languages (English and Italian). See~\cite{ludwig2016untangling} for more such efforts. 

Next, we consider real-world datasets, where the goals may be information, disruption, financial or psychological. Here we have several datasets for fake news detection~\cite{raponi2022fake},\footnote{Note that the topics can vary in a heterogeneous application, such as fake news detection, since some items could be on sport and some on politics or religion. Moreover, the goals may or may not be different. Hence, we avoid the term ``domain'' to refer to applications such as fake news.} opinion spam (aka\@ fake reviews) detection~\cite{ren2019learning}, for phishing~\cite{vermaZF19}, and a company's reward program~\cite{ludwig2016untangling}. 

Some researchers have mixed data obtained from laboratory settings with non-laboratory data, such as reviews obtained from forums. 
 For example, in~\cite{Hernandez-CastanedaCG17}, researchers analyzed three datasets: a two-class, balanced-ratio dataset of 236 Amazon reviews, a hotel opinion spam dataset consisting of 400 fabricated opinions from AMT plus 400 reviews from TripAdvisor (likely to be truthful), and 200 essays from~\cite{perez2014cross}. In~\cite{XarhoulacosASG21}, researchers studied a masking technique on two datasets: a hotel, restaurant, and doctor opinion spam dataset and the dataset from~\cite{perez2014cross}. 
In~\cite{cagnina2017detecting}, in-domain experiments were done with a positive and negative hotel opinion spam dataset, and cross-domain experiments were conducted with the hotel, restaurant, and doctor opinion spam dataset.

A few works have developed domain-independent deception datasets in our sense, wherein the goals of deception can be quite different.
In~\cite{Rill-GarciaPRE18}, researchers used two datasets: the American English subset consisting of a balanced-ratio 600 essays and transcriptions of 121 trial videos (60 truthful and 61 deceptive), which we call Real-Life\_Trial.  
In~\cite{vogler2020using}, three datasets were used: positive and negative hotel reviews, essays on emotionally-charged topics, and personal interview questions. 
In~\cite{XarhoulacosASG21}, multiple fake news datasets, a COVID-19 dataset, and some micro-blogging datasets were collected and analyzed. 
In~\cite{shahriar2021domain}, researchers collected fake news, Twitter rumors, and spam datasets.
(Spam is essentially advertising. Deception is employed to fool automatic detectors rather than the human recipient of the spam. 
We focus on human targets.)
They applied their models trained on these datasets to a new COVID-19 dataset.  
In~\cite{yeh2021lying}, seven datasets were collected (Diplomacy, Mafiascum, Open-Domain, LIAR, Box of Lies, MU3D, and Real-Life\_Trial) and analyzed using LIWC categories, without claiming domain independence or cross-domain analysis. However, their datasets do involve different goals. 
LIAR, for instance, includes political lies with the goal of winning elections, whereas the lies in Real-Life\_Trial have other goals, and Diplomacy/Mafiascum are about winning online games. 
In~\cite{feng2012syntactic}, four datasets were collected: trip-advisor gold, a balanced hotel reviews dataset of 800 reviews introduced in~\cite{ott2011finding}, trip-advisor heuristic, another balanced reviews dataset of 800 reviews collected by the authors, a third 800 review Yelp dataset of uncertain ground-truth collected by the authors, and the 296 essays on three topics dataset of~\cite{mihalceaS09}. They show that features based on CFG parse trees along with unigrams performed the best on these datasets.

Thus, we still lack large, comprehensive datasets for deception that have a wide variety of deceptive goals. 

\subsection{Detection}

Deception detection in general is a useful and challenging open problem. There have been many attempts at specific applications such as phishing and fake news. On phishing alone (query: phish), there are more than 1,700 DBLP results, including over 15 surveys and reviews. Similarly, there are over 900 papers on scams (query: scam, not all of them are relevant, since many occurrences are part of acronyms such as SCAMP), over 100 on opinion spam, close to 100 on fake reviews, and over 1,800 on fake news.\footnote{All these DBLP search results are as of 31 August 2023.} 

A soft domain transfer method is proposed in~\cite{sadatMG22}. They found that partial training on tweets helped in phishing and fake news detection. In~\cite{panda22,PandaL23}, the authors study deception detection across languages and modalities. 
Other works on domain-independent deception detection have been discussed above under Datasets.


\subsection{Reviews on Linguistic Markers}
Recently,   Gröndahl and   Asokan~\citeyearpar{grondahlA19} conducted a survey of the literature on deception. They defined implicit and explicit deception, focused on automatic deception detection using input texts, and then proceeded to review 17  papers on \textit{linguistic} deception detection techniques.
(Explicit deception is when the deceiver explicitly mentions the false proposition  in the deceptive communication.)
These papers covered two forms of deception: (a) dyadic pairs in the laboratory, where one person sends a short essay or message to another (some truthful and some lies), and  (b) fake reviews (a.k.a.\@ opinion spam).  
%
Based on their analysis of the literature on laboratory deception experiments and the literature on opinion spam, they concluded that \textit{there is no linguistic or stylistic trace that works for deception in general}. 
Similarly, the authors of~\citet{vogler2020using} assert that extensive psychology research  shows that ``a generalized linguistic cue to deception is unlikely to exist.'' 

We collectively refer to~\citet{grondahlA19}, \citet{fitzpatrickBF15}, and \citet{vogler2020using,vrij08} as 
the \textit{Critiques}.
We argue that, at best, their analyses and conclusion may be a bit too hasty and elaborate on several aspects that need investigation/analysis with specific examples from the reviewed literature. 

Although we focus on those specific critiques here, many of the issues we raise are more generally applicable to any systematic review of scientific literature. 

\section{Guidelines for Systematic Reviews} \label{sec-guide}
According to Staples and Niazi~\citeyearpar{staplesN07}, ``A systematic review is a defined and methodical way of identifying, assessing, and analyzing published primary studies in order to investigate a specific research question.'' Unlike an ad-hoc literature review, systematic reviews are formally planned and methodically executed.
Such a review can reveal the structure and patterns of existing research, highlight key results, and identify gaps for future research. 

Unsurprisingly, there has been an explosion of systematic reviews on all kinds of problems in natural language processing and machine learning. However, there is a dearth of good guidelines and procedures for them in NLP. In this section, we lay out guidelines for a good systematic review and identify several common pitfalls a team can stumble into.  



\subsection{What Makes a Good Systematic Review?}
A good systematic review is independently replicable and thus has additional scientific value over that of a literature survey.  In collecting, evaluating, and documenting all available evidence on a specific research question, a systematic review may provide a greater level of validity in its findings than might be possible in any individual study reviewed. 
However, systematic reviews require much more effort than ordinary literature surveys.

The following features differentiate a systematic review
from a conventional one (Kitchenham, 2004):
\begin{itemize}
\item[(1)] A predefined and documented protocol specifying the research question and procedures to be used in performing the review.
\item[(2)] A defined and documented search strategy designed to find \textit{as much of the relevant literature as possible}.
\item[(3)] Explicitly predefined criteria for determining whether to include or exclude a candidate study. 
\item[(4)] Description of quality assessment mechanisms to evaluate each study.
\item[(5)] Description of review and cross-checking processes involving multiple independent researchers, to control researcher bias.
\end{itemize}

\subsection{Common Pitfalls}
We identify several challenges faced by reviews and surveys, systematic or conventional. 

\paragraph{Inadequate search strategies}
Not having a clear, explicit search strategy for literature or clearly defined inclusion and exclusion criteria can lead to bias in the selection of papers.

\paragraph{Confirmation nias}
The search strategy should be designed to avoid favoring one hypothesis over another. 

\paragraph{Publication bias}
Even a rigorous and thorough search of the published literature might not give a full picture of the state of the field. Factors besides the quality of the work can influence whether a paper gets published. For example, studies with positive results, papers with well-written English, and papers authored by highly reputed researchers are more likely to get published. Longer works such as theses are also missed in the emphasis on published literature. 

\paragraph{Clique bias}
Cliques of interconnected researchers and papers analyzing the same dataset may share biases. When performing a systematic review, researchers must ensure that they are not lending too much weight to one cluster of connected works. 

\paragraph{Quality of studies and datasets}
Within the literature, there is a significant range in the quality of the studies. Quality assessment criteria should consider: (i) the design of the experiments, (ii) the sizes and the heterogeneity of the populations, (iii)   whether the statistical tests used are appropriate for the datasets analyzed, whether tests of statistical significance were applied, and correctly reported so that effect sizes can be obtained, (iv) whether something like the Bonferroni-Holm correction was used for the multiple comparisons issue, and (v) their replicability.

\section{A Critique of the Critiques} \label{sec-debate}
We now take our guidelines and apply them to the critiques. 

The deception survey of~\citet{grondahlA19} has some of the features of a good systematic review: they specify the research questions and hypotheses and involve two researchers (presumably mentor and mentee). 
However, they lack a formal review protocol, search strategy, or explicit inclusion/exclusion criteria, and no quality assessment mechanism is specified. 
However, because it is published in an influential journal, it is likely to leave a lasting impression on deception researchers, so it is worth the time and effort to examine its strengths and weaknesses. 


To check the completeness of their search, we searched the literature for relevant papers published before 2019. 
The deception survey~\cite{grondahlA19} suffers from an incomplete search. Although their goal was to survey automatic linguistic deception detection literature, they missed many relevant papers including the meta-analysis by Hauch~\citeyearpar{hauch16}.

This meta-analysis examined 79 cues from 44 different studies on automatic linguistic deception detection. They state: ``The meta-analyses demonstrated that relative to truth-tellers, liars experienced greater cognitive load, expressed more negative emotions, distanced themselves more from events, expressed fewer sensory–perceptual words, and referred less often to cognitive processes. However, liars were not more uncertain than truth-tellers. These effects were moderated by event type, involvement, emotional valence, intensity of interaction, motivation, and other moderators. Although the overall effect size was small, theory-driven predictions for certain cues received support.'' 

To check for publication bias, we performed a systematic search of the ProQuest Global Database. 
We identified 118 dissertations and theses with the keywords ``deception'' \textit{and}  ``detection'' in the title. 
Three of these also had the word ``linguistic'' in the title and all three were relevant.
Replacing ``linguistic'' with ``natural language processing'' (or ``textual'') and keeping ``deception'' yielded two more relevant dissertations. 
Finally, ``verbal'' with ``deception'' yielded four more relevant results, out of nine total. 
None of the above dissertations are cited in the Critiques. 



\begin{figure}[tp]
\centering
\includegraphics[width=0.6\columnwidth]{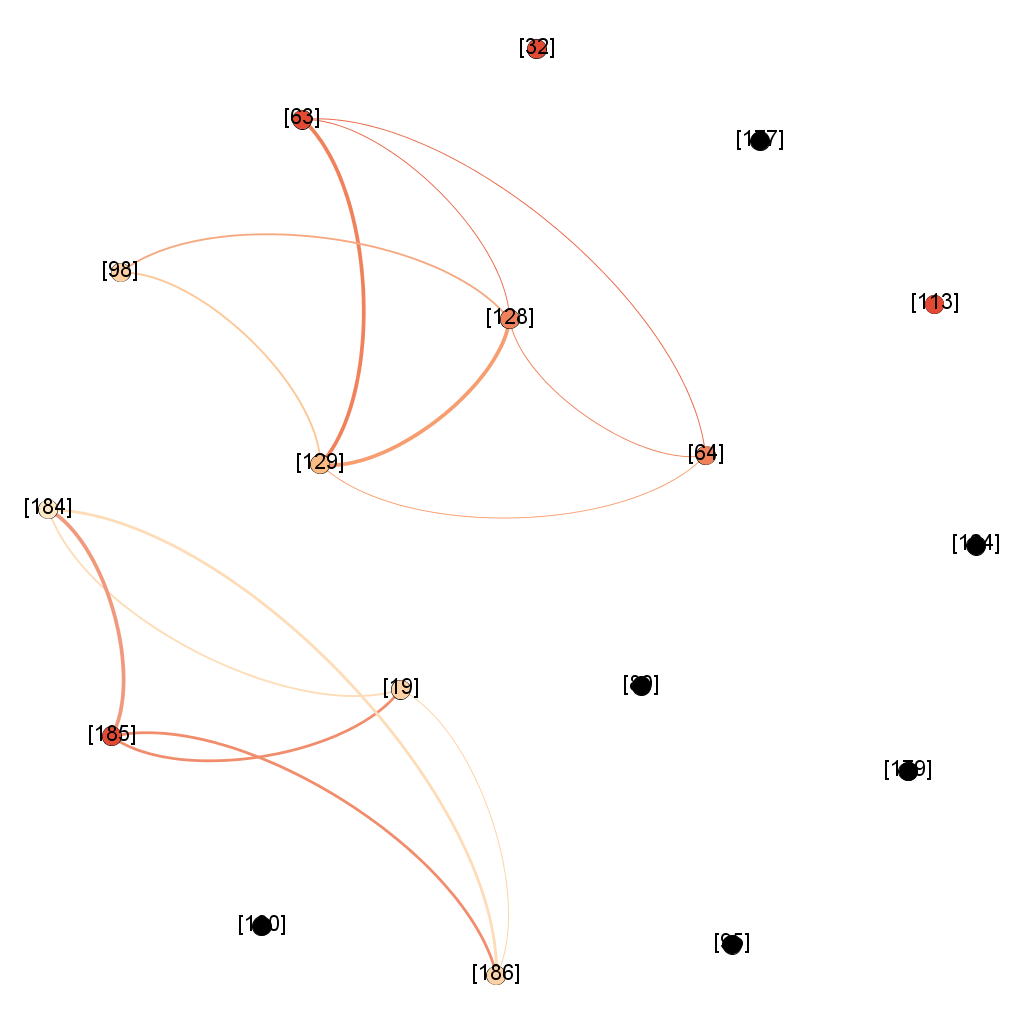}
\caption{Graph showing the 17 papers as vertices. There is an edge between papers that have a common author. The thickness and color reflect the number of common authors. $K_3$ and $K_4$ cliques are visible.} \label{fig-graph}
\end{figure}

To check for clique bias in the deception survey~\cite{grondahlA19}, we listed all the authors of the 17 papers cited in Section 2 (``Deception Detection Via Text Analysis'') of their paper and generated a graph of connected papers in Figure~\ref{fig-graph}. We consider two papers connected if they share a common author. 
We find that nine authors account for 15 (88\%) of the papers, and the connected papers graph contains several cliques, two as large as $K_4$ and one additional $K_3$. 

None of the papers examined in the meta-analysis conducted by~\cite{depauloLM03} and the deception survey in~\cite{grondahlA19} built a general dataset for different deception goals ({e.g.}, as in phishing, fake news, \textit{and} crime reports). 
If researchers study a particular form of deception and build a dataset to study it, the chance that they would stumble upon general linguistic cues for deception is likely to be small, since that was not even their objective anyway! 
Hence, a review of these papers is also unlikely to find any general linguistic cues for deception. 


\subsubsection*{Datedness}
The meta-analysis of~\cite{depauloLM03} was conducted in 2003. 
The meta-analysis of~\cite{hauch16} is more recent, but still only covers papers up to February 2012. 
The latest review of meta-analyses~\cite{sternglanzMM19} on deception detection lists more than 50 meta-analyses.
Of course, not all are relevant to linguistic deception detection, but this points to the large volume of work in the field and is indirect evidence for the contemporary inadequacy of the literature cited in the  Critiques. 


\section{Domain-Independent Markers} \label{sec-args}
Contrary to the assertion in the Critiques, there are several arguments in favor of general linguistic markers for deception.

\subsection{Prior Analyses}

The meta-analyses~\citet{depauloLM03} and~\citet{hauch16} did find small markers of deception in the studies they examined despite analyzing studies of specific forms or situations of deception, not general domain-independent datasets. 

Likewise, the following papers all point to evidence for cross-domain deception detection:~\cite{Rill-GarciaPRE18,shahriar2021domain,vogler2020using,XarhoulacosASG21,yeh2021lying}. These researchers created domain-independent datasets and developed features and techniques for deception detection across domains. 

\begin{table}[tp]
    \centering 
    \caption{The latest surveys, reviews, and meta-analysis on automatic deception detection. Are the queries (\textbf{QL}) or databases \textbf{DB} listed? \textbf{Period} of the searches. The number of \textbf{papers}  surveyed. Is there support for \textbf{ling}uistic features?}
    \label{tab:the_table}
    \begin{tabular}{|c|c|c|c|c|c|}
        \hline
      \bf  Reference &  \bf QL  &  \bf DB & \bf Period & \bf Papers &  \bf Ling? \\\hline
         GA19~\cite{grondahlA19} & No & No & -- & 18 & No \\ \hline
        H16~\cite{hauch16} & Yes & Yes & 2011--12 & 44 & Yes \\ \hline
  E19~\cite{elhadad2019fake} & Yes & Yes & 2017--19 & 47 & Partial \\ \hline
    \end{tabular}
\end{table}

The meta-analysis of~\cite{hauch16} searched four databases: PsycInfo, Social Science Citation Index, Dissertation Abstracts, and Google Scholar for articles between 1945 and February 2012 with ``all permutations and combination of one keyword from three different clusters: (i) verb, language and linguistic; (ii) computer, artificial, software and automatic; (iii) lie, deceit, decept*.''

The systematic review of~\cite{elhadad2019fake} searched Google Scholar for articles between 2017--2019 using 10 queries listed in their paper. Their queries are a \textit{proper} subset of the Boolean query 
\texttt{\footnotesize(fake $\vee$ false) news (identify $\vee$ detect) on (social media $\vee$ Twitter)},
which we repeated on Scholar on 11 November 2022, with a claim of over a million results. (Google counts are loose upper bounds.) Scholar displayed only the top 1000 results. The queries produced a total of fewer than 200  potentially relevant results. We summarize the pertinent characteristics of three recent reviews/surveys/meta-analyses in Table~\ref{tab:the_table}. 

\subsection{Our Analysis}

Since the meta-analysis of~\cite{hauch16} 
ended in February 2012, we searched Google Scholar, PsycInfo and Dissertations, and Abstracts Global, for the period 2013--2022 with the query: 
\begin{quote}
\texttt{\footnotesize(verbal $\vee$ language $\vee$ linguistic $\vee$ text $\vee$ lexical) $\wedge$ (computer $\vee$ artificial $\vee$ software $\vee$ automatic $\vee$ autonomous $\vee$ automated $\vee$ identify $\vee$ computational $\vee$ machine $\vee$ detect $\vee$~ tool) $\wedge$ (lie $\vee$ false $\vee$ fake $\vee$ deceit $\vee$ deception $\vee$ deceptive)}.
\end{quote}
We formed this query by appropriately combining the queries from~\cite{hauch16,elhadad2019fake}, adding keywords after scanning the initial results, and adding relevant synonyms from querying WordNet 3.1 with \texttt{\footnotesize deceit}, \texttt{\footnotesize identify}, and \texttt{\footnotesize lexical}. 
Adding \texttt{\footnotesize recognition} to the middle clause reduced the set of results by more than 100K, a flaw of Google Search. 
(We tried other synonyms, but the results seem irrelevant.)
Scholar claimed over a million results but only displayed the top thousand. A scan through them identified approximately 880 as potentially relevant. PsycInfo gave us around 450 matches and the Dissertations database yielded approximately 140 matches. 
The Scholar query: 
\begin{quote}
\texttt{\footnotesize(verbal $\vee$ language $\vee$ linguistic $\vee$ text $\vee$ lexical) $\wedge$ (computer $\vee$ artificial $\vee$ software $\vee$ automatic $\vee$ autonomous $\vee$ automated $\vee$ identify $\vee$ computational $\vee$ machine $\vee$ detect $\vee$~ tool $\vee$ recognize $\vee$ recognition $\vee$ recognizing) $\wedge$ (rumor $\vee$ hoax $\vee$ misinformation $\vee$ disinformation)} 
\end{quote}
over all time periods
 claimed 350K results; the top 1000 gave around 190 potentially relevant ones. 
The results were examined for feature selection and feature ranking papers.
More than one recent survey mentioned $n$-grams of part of speech tags and semantic features as examples of generalizable features. However, this analysis also revealed a lack of feature rankings for large, diverse, general datasets.








\subsection{New Developments in NLP}
Moreover,
computer science, machine learning, and NLP have come a long way in the intervening years. 
Recent breakthroughs such as attention, transformers, and pre-trained language models like BERT, have revolutionized NLP. Even if some of the previous criticisms were valid, we must reexamine the conclusions of the Critiques considering these new advances.

\section{Linguistic Cues/Analysis}\label{sec-cues}

Because of the problems enumerated above, we collect and analyze datasets for domain-independent linguistic cues to tackle: (1) the ground truth problem for deception detection, and (2) evidence of linguistic cues for deception across domains. 

A \textbf{ground truth} is something that is known to be correct, but this information is difficult to obtain, so we need models that do not rely on having too much ground truth data. Our approach is to focus on using linguistic information from the text. For the second challenge, we try to find universal linguistic markers for deception by looking for features that behave similarly across domains. We hope that an ML model built with these features could generalize across domains~\cite{gokhman}.

\subsection{Datasets}
We summarize our deception domains and scenarios below. We focus on real-world datasets.

In the \textit{product review} domain, we use the Amazon reviews dataset mentioned above~\cite{garcia}.  

In the \textit{job scam} domain, we identify fraudulent job listings. Our dataset contains the bodies of 13,735 legitimate and 608 fraudulent job listings. 

In the \textit{phishing} domain, we distinguish between legitimate emails and phishing emails. Our dataset contains the bodies of 9,202 legitimate and 6,134 phishing samples. The IWSPA-AP dataset analyzed above is a subset of this dataset. 

In the \textit{political statement} domain, we determine the truthfulness of claims made by US political speakers. Our dataset contains 7,167 truthful and 5,669 deceptive statements evaluated by PolitiFact.

In the \textit{fake news} scenario, we distinguish between legitimate and fake news. Here we use the WELFake dataset~\cite{verma}.  

We analyzed each dataset for any artifacts of data collection and cleaned them to remove such artifacts. The cleaning procedures include two parts: text removal and text cleaning.  We then sanitize the texts using the methods discussed in~\cite{Zeng_codaspy}. We remove meta-data in emails and source leaks in news and replace HTML break tags with new lines. Additionally, the authors of~\cite{Zeng_codaspy} found that the provided labels in WELFake~\cite{verma} are flipped, so we flip its labels as a final cleaning step. We are making the combined, cleaned dataset available on Zenodo at \url{https://zenodo.org/record/6512468#.ZBVRUhTMLQM}. 

\subsection{Sources for Linguistic Cues}
Function words (FW) are words that express a grammatical relationship between words in a sentence. Unlike content words, function words such as `when,' `at,' and `the' are independent of specific domains. Function words and $n$-grams are useful for many text classification tasks, including author gender classification, authorship attribution  \cite{Argamon}, and deception detection \cite{siagian}. To gain an insight into the transfer of knowledge between domains, we utilized three types of explainable features: function words, part-of-speech (POS) tags of function words, and engineered linguistic features. POS tags were used to determine whether a word was a function or a content word; the content words were then removed. The last experiment utilized 151 engineered linguistic features ($13+55+86 - 3$ duplicates removed by the colinearity check below). 

The engineered features are drawn from three sources. Linguistic Inquiry and Word Count~\cite{boyd2022development}, a popular source of features in the NLP literature, was the source of 86 features. The authorship attribution paper~\cite{fabien2020bertaa}  was the source of 55 features.  Thirteen features were collected from two papers, one on deception~\cite{zhou} and the other on fake news~\cite{verma}, after significance testing using $t$-tests with and without the Bonferroni-Holm correction of $p$-values. 

The initial significance testing of 27 linguistic features from the two papers~\cite{zhou,verma} on four public datasets is described in  Appendix A. Appendix B describes an analysis of function word $n$-grams on the same datasets as in Appendix A. 
A complete source-wise list of the 55 features from~\cite{fabien2020bertaa} and 86 features from~\cite{boyd2022development} is in Appendix C. 
Function words as features for deception have been studied before,  in~\cite{siagian}, for example. We also experimented with the part-of-speech tags of function words.

%



\subsection{Results of Feature Analysis}

\begin{table}[tp]
    \caption{Function word features: N  -- number of common features; F -- fake news, J -- job scams; P -- phishing; Pr -- product reviews, Ps -- political statements; CC -- cumulative count including common features inherited from supersets.}
    \centering
    \begin{tabular}{|c|c|c|c|}
    \hline
        Subset & N & Common Features & CC \\ \hline
        All & 6 & and, in, is, of, on, the & 6 \\ \hline
        F, J, P, Pr & 2 & this, you & 8  \\ \hline
        F, J, Pr, Ps & 1 & are & 7 \\ \hline
        J, P, Pr, Ps & 2 & for, to & 8  \\ \hline
        F, J, P & 1 & at & 9 \\ \hline
        F, Pr, Ps & 3 & it, that, would & 10 \\ \hline
        J, P, Ps & 2 & from, our & 10 \\ \hline
        J, Pr, Ps & 2 & as, with & 11 \\ \hline
         P, Pr, Ps & 1 & not & 9 \\ \hline
        F, P & 2 & all, had & 11 \\ \hline
        F, Ps & 1 & he & 11 \\ \hline
        J, Pr & 2 & be, or & 13 \\ \hline
        J, Ps & 1 & we & 12 \\ \hline
        P, Pr & 1 & me & 10 \\ \hline
        Pr, Ps & 2 & they, was & 17 \\ \hline
    \end{tabular}
    \label{tab:fw_features}
\end{table}

\begin{figure}[tp]
\centering
\subfloat[Accuracies]{
\includegraphics[width=0.45\linewidth]{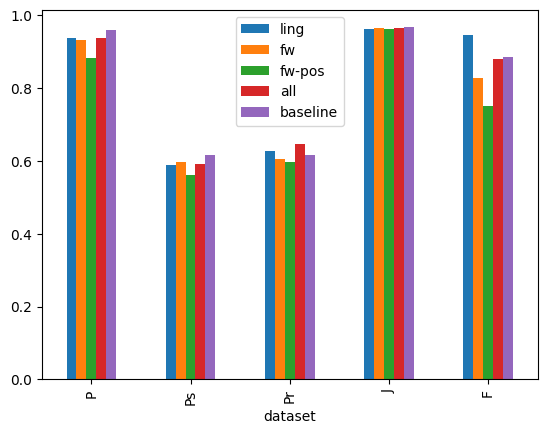}
}
\subfloat[F1-scores]{
\includegraphics[width=0.45\linewidth]{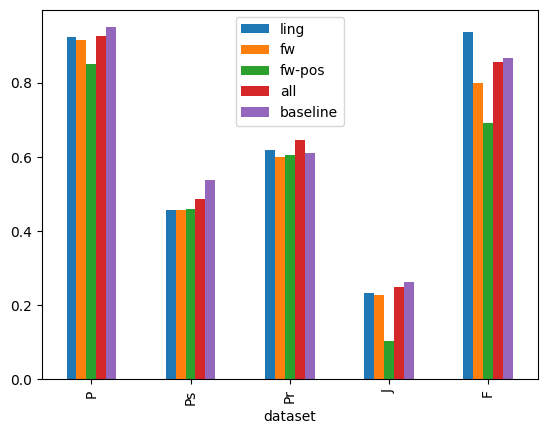}
}
\caption{Random Forest performance for the five feature types: linguistic (ling), function words (fw), pos tags of function words (fw-pos), combination of the three (all), and unigram tfidf  (baseline); F -- fake news; J -- job scams; P -- phishing; Pr -- product reviews; Ps -- political statements.}
\label{fig:accuracies}
\end{figure}

\begin{table}[tp]
    \centering
        \caption{Function word part-of-speech features: N -- number of common features; F -- fake news; J -- job scams; P -- phishing; Pr -- product reviews; Ps -- political statements; CC -- cumulative count including common features inherited from supersets. }
    \begin{tabular}{|c|c|c|c|}
    \hline
        Subset & N & Common Features & CC \\ \hline
         All & 10 & CC, CD, DT, IN, MD, PRP, RB, TO, VBP, VBZ & 10 \\ \hline
         F, J, P, Ps & 5 & RP, VB, WDT, WP, WRB & 15 \\ \hline
         F, P, Pr, Ps & 1 & VBD & 11 \\ \hline
         F, J, P & 2 & POS, UH & 17 \\ \hline
         F, P, Ps & 2 & EX, VBN & 18 \\ \hline
         F, J & 1 & VBG & 18 \\ \hline
         J, P & 1 & ADD & 18 \\ \hline
    \end{tabular}
    \label{tab:fw_pos_features}
\end{table}

\begin{table}[tp]
    \centering
        \caption{Engineered Linguistic Features: N  -- number of common features; F -- fake news; J -- job scams; P -- phishing; Pr -- product reviews; Ps -- political statements;  CC -- cumulative count including common features inherited from supersets.}
    \begin{tabular}{|c|c|c|c|}
    \hline
        Subset & N & Common Features & CC\\ \hline
         All & 1 & per\_cap & 1 \\ \hline
         J, P, Pr, Ps & 5 & Dic, f\_b, f\_g, per\_digit, richness & 6 \\ \hline
         F, J, P & 4 & cert, f\_e\_2, function, sen\_len & 5 \\ \hline
         F, J, Pr &  1 & period & 2 \\ 
         \hline
         F, P, Pr & 1 & paus & 2 \\ 
         \hline
         P, Pr, Ps & 3 & conj, f\_f, modi & 9 \\ \hline
         F, J & 2 & apostro, comm &  8 \\ \hline
         F, P & 5 & f\_e\_0, f\_e\_1, f\_e\_3, f\_e\_7, sens & 11 \\ \hline
         F, Pr & 10 & adverb, allPunc, analytic, f\_e\_8, focuspast, & 13 \\
         & & ipron, len\_text, OtherP, pronoun, WPS & \\ 
         \hline
         J, P & 5 & f\_c, f\_o, f\_v, f\_w, socrefs & 15 \\ \hline
         P, Pr & 7 & avg\_len, f\_d, f\_i, f\_s, f\_t, f\_y, self\_ref & 17 \\ \hline
         P, Ps & 2 & f\_1, f\_p & 11 \\ \hline
         J, Pr & 3 & allure, article, lifestyle & 10 \\ \hline
         Pr, Ps & 1 & quantity & 10 \\ \hline
    \end{tabular}
    \label{tab:ling_features}
\end{table}

We used the Stanza~\cite{Stanza} POS tagger and OntoNotes Release 5.0/Penn Treebank~\cite{10.5555/972470.972475} tagset in all experiments involving POS tags. This tagset builds on top of the original Penn Treebank, and adds seven new tags: 

\begin{quote}
\texttt{ADD - Email, AFX - Affix, HYPH - Hyphen, NFP - Superfluous punctuation, UH - 
 Interjection, SP - Space, and XX -  Unknown.}
\end{quote}

Due to the parser's limitations, several samples of text that had a length more than one million characters had to be discarded. We did not remove stop words or further alter the data in any manner. Function words and their respective POS tags were separately vectorized as word unigrams using the tf-idf scheme. The raw texts were processed and vectorized identically and used as a baseline. The motivation behind it was to (i) understand whether it is possible to achieve similar results while using only a few non-domain-specific features that are highly indicative of deception, and (ii) investigate the impact of content words on deception through the contrast between the baseline and function words.

For each dataset, and for each set of features, we applied three techniques to select the most relevant features. First, a random forest algorithm~\cite{breiman2001random} was used, which allowed us to rank features by their importance. The least important ones were removed under the condition that the out-of-bag accuracy on the validation set either increased or remained the same after removing the features. Next, we applied scipy's~\cite{2020SciPy-NMeth} single linkage hierarchical clustering~\cite{Gower1969MinimumST} with Spearman's correlation~\cite{spearman1961proof} as the measure of feature colinearity. Features exhibiting a high degree of colinearity were removed with their redundancy validated in the same manner as with the first technique. Finally, taking the remaining features, we applied Hyperopt's~\cite{bergstra2013hyperopt} feature selection and the eXtreme Gradient Boosting algorithm~\cite{Chen:2016:XST:2939672.2939785} with SHAP~\cite{NIPS2017_7062} as a metric of each feature's contribution to the overall model performance. Ultimately, the aforementioned approach produced a subset of the features for each of the five datasets. A total of $81$ linguistic, $28$ function word POS, and $61$ function word features were selected; $50/81$, $22/28$, and $29/61$ were shared with at least one other dataset.

For our analysis of the potential for knowledge transfer any feature unique to a dataset was removed, leaving only those significant for at least two datasets and therefore being of interest for understanding of transfer between domains. The relationships of function words and function words' POS tags across datasets are depicted in Tables~\ref{tab:fw_features} and~\ref{tab:fw_pos_features}, while linguistic feature transfer is summarized in Table~\ref{tab:ling_features}. The cumulative count (CC) in these tables serves as a measure of how many features a group of domains has in common. Several trends can be noticed from these three tables. For example, all five datasets share $6+10+1 = 17$ common features, and the fake news, job scams, and phishing datasets have a total of $9+17+5 = 31$ features in common. Also, the subset $\{$F, J$\}$ has $9+18+8=35$ common features, and $\{$J, P, Pr, Ps$\}$ has $8+10+8 = 26$ common features. Job scams and phishing together have 43 common features. Similarly, we can see that deceptive attacks can be differentiated using  features such as `to', personal pronouns, singular present verb forms, modals, and adverbs (compare with the quote from Rowe~\cite{rowe2006taxonomy} in Section~\ref{sec-strata}). The richness, possessive ending and interjection features are significant for fake news, job scams and phishing. Fake news and product reviews have many significant LIWC features. 

Datasets that share a significant number of features are good candidates for domain adaptation; however, the performance of a model using a potentially limited set of features shared across tasks must remain robust. To this end, we combined previously selected linguistic, function words, and function word POS features that were shared by two or more datasets. This resulted in a final set of $91$ features. Upon further applying feature selection, the number of significant features of all three types shared among datasets has been reduced to $45$.

To evaluate the features' performance,  we used a random forest classifier with five-fold cross-validation. The model hyperparameters were set to $50$ trees with the leaf nodes of $5$ samples, and $50\%$ of the features were considered on each split. Gini impurity was used as a criterion of split quality. 

The accuracy and F1-scores of the model using each of the feature sets across the five datasets are shown in Figure~\ref{fig:accuracies}. It is important to note that Job Scams' data appears to be heavily imbalanced and the models' performance on it is not an ideal indicator of feature quality. Generally, the combined set of shared features is nearly on par with the baseline, with linguistic, function word, and function word POS following in the order given. Notable exceptions are Product Reviews where linguistic and combined features beat the others, including the baseline, and Fake News with linguistic features outperforming the rest by a significant margin. We hypothesize that the relative length and richness of news articles may be in part responsible for this phenomenon.

\section{Deep Learning Based Experiments} \label{sec-dl}  
To investigate the possible existence of other deception signals, we turn to deep learning. If universal deception signals exist, then a deep-learning model can learn to detect them. To determine whether this happens, we perform two experiments on the same five cleaned datasets of the previous section.
First, we evaluate the performance of models trained on multiple domains. Then we train models on one domain and evaluate their performance on other domains.

\subsection{Model}
Our model architecture consists of a base pre-trained transformer model, a dropout layer, and a linear layer. As standard in NLP, we prepend a [CLS] token to the text, pass the text through the base model, and perform classification on the last-layer embedding of the [CLS] token. 

\begin{ignore}\centering
    \includegraphics[scale=0.5]{crossdataset_dl/Model.pdf}
    \caption{Model used for deep learning experiments.}
\end{ignore}

\subsection{Multi-domain Experiment}
If deep-learning models trained on multiple domains pick up on universal deception signals, then we should expect performance on \emph{individual} domains to be positively correlated amongst each other. Conversely, if they only learn domain-specific signals, then we should expect performance on individual domains to be negatively correlated with one another.

We train 100 models on the union of our datasets. We use a random 80/10/10 train/validate/test split for each dataset with uniformly drawn hyperparameters. We use BERT-base and RoBERTa-base for our base models, dropout percentages between 0.1 and 0.5, and the AdamW optimizer with learning rates between $0.00001$ and $0.0001$.

We then evaluate each model on the individual test sets. We exclude models that failed to converge and models that have an outlier F1 score using the IQR test and perform pairwise linear regression on the remaining F1 scores. 

\begin{sidewaysfigure}[p]
    \includegraphics[width=7 in]{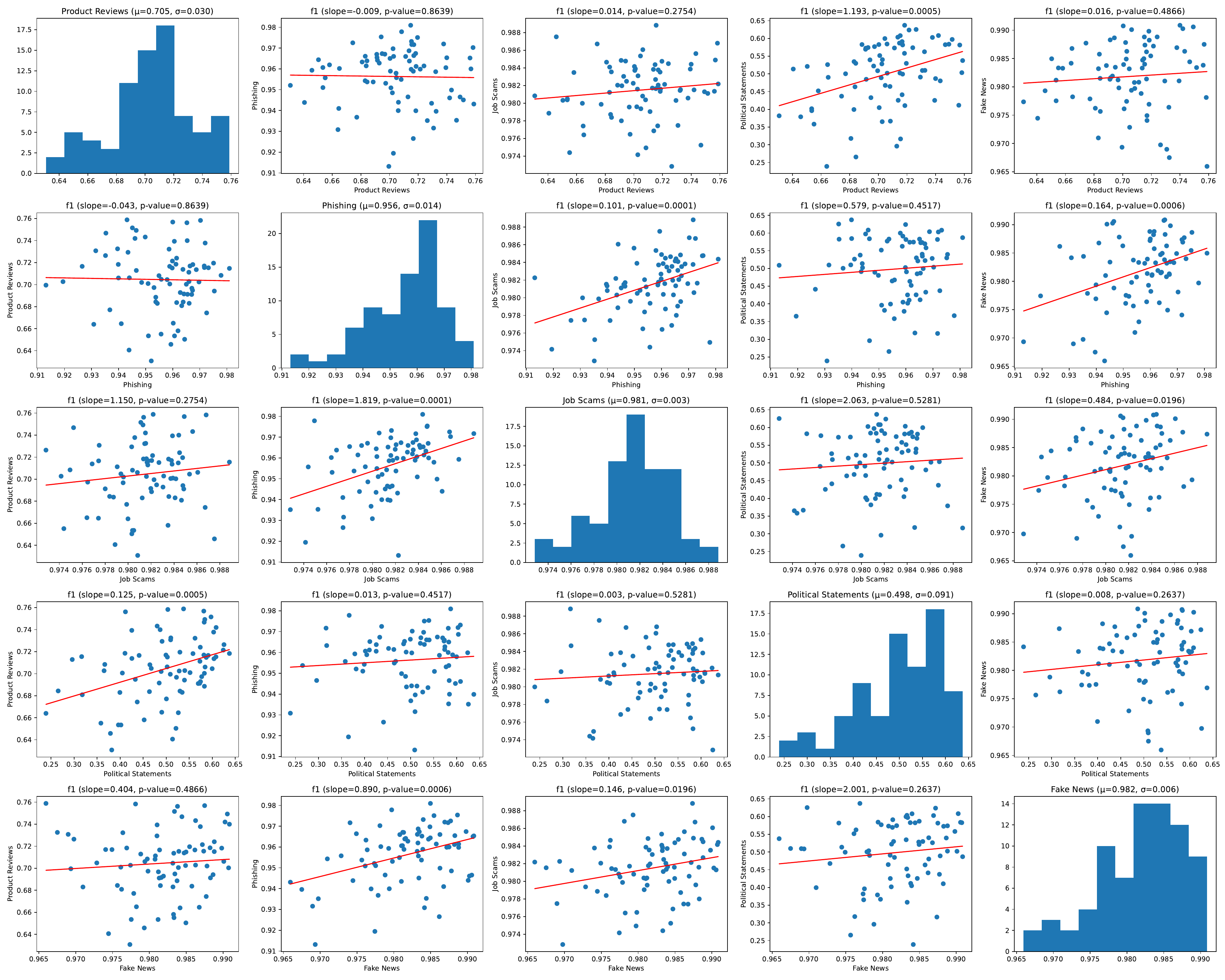}
    \caption{Pairwise F1 score scatter matrix of converged combined models. Outliers are excluded.}
    \label{fig:cross_dataset_scatter_no_outlier}
\end{sidewaysfigure}

We present our results without outliers in Figure~\ref{fig:cross_dataset_scatter_no_outlier}. All pairs of tasks except for product reviews and phishing are positively correlated, with five of them significant at the 0.05 level. 


\subsection{Cross-Domain Generalization Experiment}
If a deep-learning model primarily learns a universal deception signal, then it should generalize to deception domains that it has not yet seen. In particular, they should be able to achieve a higher F1 score than a coin flip classifier, which we can calculate using the formula
$\text{CF F1} = {q}/({0.5+q})$,
where $q$ is the portion of the dataset that is deceptive. 

On each dataset, we train 100 models with hyperparameters drawn from uniform distributions. We use BERT-base and RoBERTa-base for our base models and values between 0.1 and 0.5 for dropout percentage. For our learning rate, we use a different range for each task to minimize divergence: 
$[0.00001, 0.00006]$ for product reviews, $[0.00001, 0.000025]$ for job scams, $[0.00001, 0.00010]$ for phishing, $[0.00001, 0.00004]$ for political statements, and $[0.00001, 0.00010]$ for fake news. 

Each model is evaluated on each dataset, ignoring models that fail to converge.
We perform a 1-sample $t$-test with the alternative hypothesis ``the mean F1 in domain Y of models trained on X is less than or equal to the coin flip F1 of Y.''  
We report the resulting $p$-values in Table~\ref{tab:cross_dataset_p_value}. 
In ten cases, models trained on one domain manage to beat the coin-flip baseline at a 0.01 significance level, with nine cases beating the coin-flip baseline at the $10\sigma$ ($p< 7.62\times10^{-24}$) level. 
However, we also find that eight pairs have a $p$-value of 1.00, meaning they performed worse than the coin-flip baseline. 

Interestingly, we also find that the fake news models manage to beat the coin flip on all domains. We suspect that this is due in part to its larger size but leave this as a direction for future research. 

\begin{table*}
    \centering
    \caption{$p$-values in the Cross Dataset Experiment. Values below 0.01 are considered significant. A dagger indicates that the 0.00 values are correct to two decimal places. The 0.00 values without the dagger have a 0  in at least the third place after the decimal.}
    \label{tab:cross_dataset_p_value}
    \begin{tabular}{|c|lllll|}
    \hline
    \textbf{}            & \multicolumn{1}{c}{Product  } & \multicolumn{1}{c}{} & \multicolumn{1}{c}{Job } & \multicolumn{1}{c}{Political } & \multicolumn{1}{c|}{Fake } \\
  &  \multicolumn{1}{c}{  Reviews} & \multicolumn{1}{c}{Phishing} & \multicolumn{1}{c}{ Scams} & \multicolumn{1}{c}{ Statements} & \multicolumn{1}{c|}{ News} \\\hline
    Product   Reviews    & \textbf{$0.00^\dagger$}                   & 1.00                   & $0.00^\dagger$                      & 1.00                                & 1.00                      \\
    Phishing             & 1.00                            & \textbf{$0.00^\dagger$}          & $0.00^\dagger$                      & 1.00                                & 1.00                      \\
    Job Scams            & $0.00^\dagger$                           & 0.98                    & \textbf{$0.00^\dagger$}            & $0.00^\dagger$                               & $0.00^\dagger$                     \\
    Political Statements & 1.00                            & 1.00                    & $0.00$                      & \textbf{$0.00$}                       & 0.96                      \\
    Fake News            & $0.00^\dagger$                           & $0.00^\dagger$                    & $0.00^\dagger$                      & $0.00^\dagger$                                & \textbf{$0.00^\dagger$}             \\ \hline
    \end{tabular}
\end{table*}

\subsection{Discussion}
    The multi-domain experiment provides strong support for the existence of universal deception signals. All but one pair are positively correlated. Five are statistically significant, and the one negative correlation is not statistically significant. In contrast, the results of our cross-domain generalization experiment are mixed. While some pairs beat the coin-flip baseline, others performed worse than the baseline.
    
    Taken together, these results suggest that both universal and domain-specific deception signals exist. Models trained on a single task will learn both universal and task-specific signals, potentially resulting in poor generalization to other deception domains. Therefore, training a domain-independent deception detector will likely require a diverse domain-independent dataset. 
    
\section{Conclusions} \label{sec-concl}
We have provided new definitions for deception based on explanations and probability theory. We gave a
new taxonomy of deception that clarifies the explicit and implicit elements of deception. We have given sound desiderata for systematic review and meta-analysis, which we hope will help researchers conduct high-quality analyses of the literature and devise new domain-independent deception detection techniques. 

We have argued against hasty conclusions regarding linguistic cues for deception detection and especially their generalizability. 
The Critiques contained in~\cite{fitzpatrickBF15,vogler2020using,vrij08} may present a valid point, namely that some linguistic cues might not generalize across the broad class of attacks. However, over-generalizations should be made with caution, as they discourage future domain-independent deception research. Moreover, we have presented evidence showing that there do exist common linguistic cues in deceptive attacks with widely varying goals and topical content.


Our linguistic analysis of four datasets and cross-dataset analysis of five different deception datasets shows that there are linguistic features, some at the surface level and some deeper,  that can be used to build classifiers for more general deception datasets. With all the new developments in machine learning and NLP, we believe that research on linguistic deception detection is poised to take off and could result in significant advances.

\begin{ignore}

\begin{figure}[tp]
    \centering
    \includegraphics[width=\columnwidth]{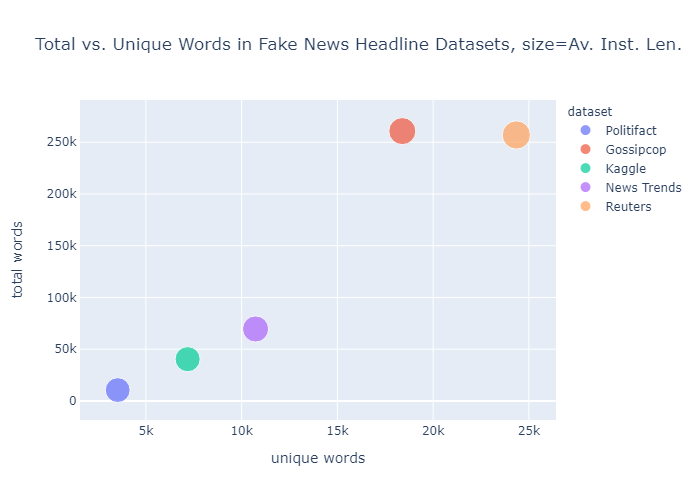}
    \caption{Unique vs. total words in headlines; size= avg. instance length.}
    \label{fig:sc:head}
The scatter plots show the vocabulary versus the total words in the datasets. 

\end{figure}
\begin{figure}[tp]
    \centering
    \includegraphics[width=\columnwidth]{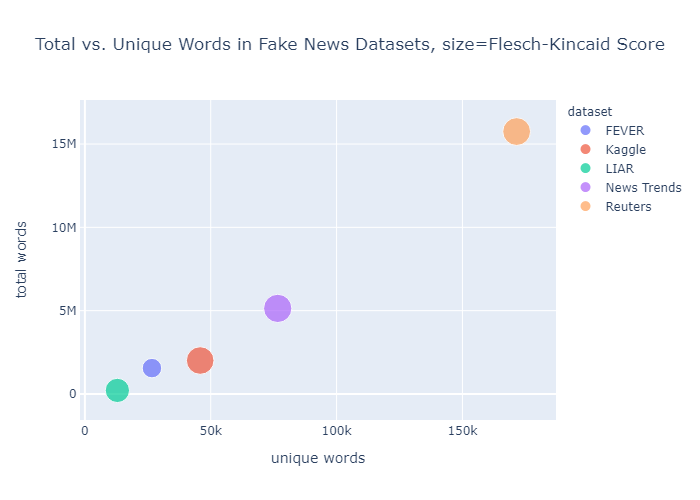}
    \caption{Unique vs.\@ total words in body text; size=FK score.}
    \label{fig:sc:body}
\end{figure}

\begin{table}[tp]\centering
\begin{tabular}{|l|l|lllllll|}
\hline
                    &  \bf D & \multicolumn{1}{l|}{ \bf TTR}                        & \multicolumn{1}{l|}{ \bf I.L.} & \multicolumn{1}{l|}{ \bf S.L.} & \multicolumn{1}{l|}{ \bf F.K.G.} & \multicolumn{1}{l|}{ \bf C.S.} & \multicolumn{1}{l|}{ \bf $n$} &  \bf POS $n$               \\ \hline
                    & F       & \cellcolor[HTML]{FD6864}{\color[HTML]{FD6864} } & \cellcolor[HTML]{FD6864}      & \cellcolor[HTML]{FD6864}     & \cellcolor[HTML]{FD6864}      & \cellcolor[HTML]{67FD9A}        & \cellcolor[HTML]{FD6864}    & \cellcolor[HTML]{FD6864} \\ \cline{2-2}
                    & K       & \cellcolor[HTML]{FD6864}                        & \cellcolor[HTML]{FD6864}      & \cellcolor[HTML]{FD6864}     & \cellcolor[HTML]{FD6864}      & \cellcolor[HTML]{FD6864}        & \cellcolor[HTML]{FD6864}    & \cellcolor[HTML]{FD6864} \\ \cline{2-2}
                    & L       & \cellcolor[HTML]{FD6864}                        & \cellcolor[HTML]{FD6864}      & \cellcolor[HTML]{FD6864}     & \cellcolor[HTML]{67FD9A}      & \cellcolor[HTML]{67FD9A}        & \cellcolor[HTML]{67FD9A}    & \cellcolor[HTML]{67FD9A} \\ \cline{2-2}
                    & NT      & \cellcolor[HTML]{FD6864}                        & \cellcolor[HTML]{FD6864}      & \cellcolor[HTML]{FD6864}     & \cellcolor[HTML]{FD6864}      & \cellcolor[HTML]{FD6864}        & \cellcolor[HTML]{FFCC67}    & \cellcolor[HTML]{FFCC67} \\ \cline{2-2}
\multirow{-5}{*}{B} & R       & \cellcolor[HTML]{FD6864}                        & \cellcolor[HTML]{FD6864}      & \cellcolor[HTML]{FD6864}     & \cellcolor[HTML]{FD6864}      & \cellcolor[HTML]{67FD9A}        & \cellcolor[HTML]{FD6864}    & \cellcolor[HTML]{FD6864} \\ \hline
                    & PF      & \cellcolor[HTML]{67FD9A}                        & \cellcolor[HTML]{FD6864}      & \cellcolor[HTML]{FD6864}     & \cellcolor[HTML]{FFFFFF}N/A   & \cellcolor[HTML]{67FD9A}        & \cellcolor[HTML]{FD6864}    & \cellcolor[HTML]{FD6864} \\ \cline{2-2}
                    & GC      & \cellcolor[HTML]{67FD9A}                        & \cellcolor[HTML]{FD6864}      & \cellcolor[HTML]{FD6864}     & \cellcolor[HTML]{FFFFFF}N/A   & \cellcolor[HTML]{67FD9A}        & \cellcolor[HTML]{FD6864}    & \cellcolor[HTML]{FD6864} \\ \cline{2-2}
                    & K       & \cellcolor[HTML]{FD6864}                        & \cellcolor[HTML]{FD6864}      & \cellcolor[HTML]{FD6864}     & \cellcolor[HTML]{FFFFFF}N/A   & \cellcolor[HTML]{FD6864}        & \cellcolor[HTML]{FD6864}    & \cellcolor[HTML]{FD6864} \\ \cline{2-2}
                    & NT      & \cellcolor[HTML]{FD6864}                        & \cellcolor[HTML]{FD6864}      & \cellcolor[HTML]{FD6864}     & \cellcolor[HTML]{FFFFFF}N/A   & \cellcolor[HTML]{FFCC67}        & \cellcolor[HTML]{FD6864}    & \cellcolor[HTML]{FD6864} \\ \cline{2-2}
\multirow{-5}{*}{H} & R       & \cellcolor[HTML]{FD6864}                        & \cellcolor[HTML]{FD6864}      & \cellcolor[HTML]{FD6864}     & \cellcolor[HTML]{FFFFFF}N/A   & \cellcolor[HTML]{67FD9A}        & \cellcolor[HTML]{FD6864}    & \cellcolor[HTML]{FD6864} \\ \hline
\end{tabular}
\caption{Dataset rankings.}
{
\begin{tablenotes}
      \item Column titles: Type-Token Ratio: TTR, Average Sentence Length (words): S L., Average Instance Length (words): I.L., Flesch-Kincaid Grade: F.K.G., Cosine Similarity: C.S., Word $n$-grams: $n$, POS $n$-grams: POS $n$. Row Titles: Headline Text: H, Body Text: B. 
      \item Color-coded based on the difference in classes, red: significant difference, yellow: some difference, green: minimal difference.
    \end{tablenotes}
    }
\label{rankingstable}
\end{table}
\end{ignore}

\begin{acknowledgments}
We thank all those who supplied datasets for this research and Vu Minh Hoang Dang for his comments on a previous draft of this article.

Verma's research was partially supported by NSF grants 1433817, 1950297, 2210198, and 2244279, ARO grants W911NF-20-1-0254, W911NF-23-1-0191, and ONR award N00014-19-S-F009.  
He is the founder of Everest Cyber Security and Analytics, Inc. 
Boumber's research was partly supported by ARO award W911NF-20-1-0254.
Zeng's research was supported by ONR award N00014-19-S-F009.
Liu's research was supported by NSF award 1950297.
\end{acknowledgments}

\bibliography{eacl,ref}

\appendix
\def\thesection{\Alph{section}}
\def\thetable{A\arabic{table}}
\def\thefigure{A\arabic{figure}}

\appendixsection{Significance Testing of Linguistic Cues from Deception Literature}\label{App:A}

For significance-testing of the linguistic cues from the deception literature~\cite{zhou2007ontology,verma}, we did a preliminary analysis of these four datasets. Dec\-Op~\cite{capuozzo} containing deceptive opinions, the WELFake dataset~\cite{verma} containing fake news, the IWSPA-AP dataset~\cite{vermaZF19} containing phishing/legitimate emails, and the Amazon Reviews dataset~\cite{garcia} consisting of truthful and fake product reviews. All are publicly available, but one of them, DecOp, is a small laboratory dataset with its limitations. 

The goals in these datasets are quite diverse. Phishing email attackers wish to install malware or steal identity/money. Deceptive opinion/review authors wish to sway opinions on services or products. Fake news authors wish to sway elections, divide people, or cause chaos. Fake product reviews are designed to sell more of a certain product or depress the sales of competitors. 

\begin{itemize}

\item {The DecOp Dataset}: 
This dataset is from \cite{capuozzo}. It contains truthful/deceptive opinions on several topics such as abortion and cannabis legalization. These opinions were collected using crowdsourcing in the US and Italy. The researchers also trained transformer models that achieved 0.62--0.90  accuracies with different settings. 

\item {The WELFake Dataset}:
This dataset is from~\cite{verma}.
It draws upon multiple true/fake news datasets.

\item {The IWSPA-AP Dataset}:
This is the IWSPA Anti-Phishing competition dataset 
of emails~\cite{vermaZF19}. 

\item {The Amazon Reviews Dataset}:
This comprises real and fake Amazon reviews from a Kaggle repository~\cite{garcia}. 
\end{itemize}

Dataset statistics are shown in Table~\ref{tab:dataset_statistics}. Of these, the WELFake and Amazon Reviews Dataset are also included in the main sections. 

We analyzed each dataset for any artifacts of data collection and cleaned them to remove such artifacts. The cleaning procedures include two parts: text removal and text cleaning.  We removed a total of 10,728 duplicate, non-English, or empty bodies, giving a total of 89,353 items. We then sanitize the texts using the methods discussed in~\cite{Zeng_codaspy}. We remove meta-data in emails and source leaks in news and replace HTML break tags with new lines. Additionally, the authors of~\cite{Zeng_codaspy} found that the provided labels in WELFake~\cite{verma} are flipped, so we flip its labels as a final cleaning step.
We removed a total of 10,728 duplicate, non-English, or empty bodies, giving a total of 89,353 items.


\begin{table}
\setlength{\tabcolsep}{4pt}
    \centering 
    \caption{Statistics of the four available datasets covering different domains.}\label{tab:dataset_statistics}
\begin{tabular}{|l|r|r@{~/~}l|l|} \hline
\bf Dataset  & \multicolumn{1}{l|}{\bf Size}   & \bf Truthful &\bf Deceptive & \bf Category           \\ \hline
Amazon Reviews    & 20,976   & 10,481 & 10,495            & Fake reviews \\ \hline
DecOp  & 1,250 & 625 & 625      & Deceptive opinions          \\ \hline
IWSPA-AP & 5,026  & 4,429 & 597           & Phishing emails    \\ \hline
WELFake   & 62,101 & 34,615 & 27,486      & Fake news \\ \hline\hline
\bf {Total} & 89,353 & 50,150 & 39,203 & Combined \\\hline
\end{tabular}
\end{table}


\subsection{Features: Linguistic Cues}


We extracted 27 total textual features from the literature shown in Table~\ref{tab:cues}. Features 1--27, with 15, 16 and 23 skipped, are from \cite{zhou}, and features 28--30 are from \cite{verma}. We then measure their values on the deceptive and legitimate samples of each dataset and count the number of datasets with statistically significant differences (using appropriate statistical tests). A difference is significant if and only if its $p$-value, after  Bonferroni-Holm correction, is smaller than the threshold  0.01. Since there is some debate on the multiple comparisons issue ({e.g.}, see \cite{gelman2012we}), we report statistically significant features both with and without the correction in Table~\ref{tab:features_short}.

\begin{sidewaystable}[p]
 \caption{List of linguistic cues. Features 1--27 are from~\protect\cite{zhou}, and features 28--30 are from~\protect\cite{verma}. Features with an asterisk ($*$) are selected after testing on the four available datasets. Features in \textbf{bold} remain qualified after adjusting their $p$-values per the Bonferroni-Holm method.} \label{tab:cues}
 \scriptsize
\begin{tabular}{|p{20mm}|p{75mm}||p{20mm}|p{75mm}|}
 \hline
1. \textbf{words*} (words) & W(D). NLTK’s word tokenizer~\cite{loper2002nltk} was used to identify words. 
& 16. generalizing terms [skipped] & {\it Missing computational description.} The deﬁnition is: refers to a person (or object) as a class of persons or objects that includes the person (or object). \\
\hline
2. \textbf{verbs*} (verbs) & Num-verbs(D). NLTK's word tokenizer was used to identify verbs. 
 & 17. \textbf{self reference*} (self\_ref) & Num-ﬁrst person singular pronouns(D) ({i.e.}, Num-$\{I, me\}/\textrm{W(D)}$. \\
 \hline
 3. noun phrase & Num-noun phrases(D). The noun chunk function in spaCy~\cite{spacy} was used to identify noun phrases. 
& 18. group reference & Num-ﬁrst person plural pronoun(D) ({i.e.}, Num-$\{we, us\}) / \textrm{W(D)}$. \\
 \hline
 4. \textbf{sentence*} (sens) & S(D). NLTK’s sentence tokenizer was used. 
& 19. emotiveness & (Num-adj.(D) + Num-adv.(D)) / (Num-nouns(D) + Num-verbs(D)). \\
\hline
5. average number of clauses & The average number of clauses per sentence. Stanza~\cite{Stanza} was used for POS tagging. Numclauses = Num-verb predicates (word.upos = `VERB') $-$ Num-root (word.deprel = `root') $-$ Num-conjugations (word.deprel = `conj'). 
 & 20. lexical diversity & Num-distinct words / W(D). \\
 \hline
6. \textbf{average sentence length*} (sen\_len) & W(D) / S(D) 
& 21. content word diversity & Num-unique content words(D) / Num-content words(D). Content words are words with lexical meanings, as opposed to function words. Methods to identify content/function words are discussed in the Function Word $n$-gram section in the Appendix. \\
 \hline
 7. average word length* (word\_len) & Num-characters(D) / W(D). Characters include digits, punctuation, and spaces. 
 & 22. redundancy* (redun) & Num-function words(D) / S(D). \\
 \hline
8. average length of noun phrase (NP) & Num-words in noun phrases(D) / Num-noun phrase(D). Noun phrases are identiﬁed in the same way as in Feature (3). 
 & 23. typographical error ratio {[}skipped{]} & \textit{This feature is skipped because exploratory analysis showed that the typographical error ratio is zero for most texts in both categories. The popularity of the auto-correct feature on browsers and text editing software has probably diminished the effectiveness of this feature.} \\
 \hline
9. \textbf{pausality*} (paus) & Num-punctuation marks(D) / S(D) 
 & 24. spatiotemporal information  & Num-( `space’ + `time’) / W(D), where `space’ and `time’ refer to the Num-words with the tag `space’ and `time’ in the LIWC2015 dictionary. LIWC is a dictionary that associates words with various tags; we used a commercial program from Pennebaker Conglomerates~\cite{pennebaker}.  \\
 \hline
 10. \textbf{modiﬁer*} (modi)  & Num-adjectives and adverbs(D).  A word is an adjective or adverb iff word.upos = `ADJ' or word.upos = `ADV'. Stanza was used for POS tagging.
 & 25. perceptual information & Num-`percep’ / W(D). `percep’ is deﬁned similarly,  as in Feature (24). \\
 \hline
 11. modal verb*  (modal) & Num-modal verbs(D) / W(D).  A word is a modal verb iff word.upos = `AUX' and word.xpos = `MD'.  Stanza was used for POS tagging.
 & 26. positive affect & Num-`posemo’ / W(D). `posemo’ is deﬁned similarly, as in Feature (24).\\
\hline
12. certainty* (cert) & Num-words that have the tag `certain’ in the LIWC2015 dictionary(D) / W(D). 
& 27. negative affect & Num-`negemo’ / W(D). `negemo’ is deﬁned similarly in Feature (24). \\
 \hline
 13. other reference  & Num-third person pronoun(D) / W(D). A word is a third person pronoun iff word.xpos = `PRP' and word.feats = `Person=3'. 
 We used Stanza for POS tagging. 
 & 28. Gunning fog grade readability index~\cite{Gunning} & An index to quantify the readability of a text by estimating the years of education required to understand the text. We used TextSTAT~\cite{textstatcite} to calculate it. \\
 \hline
14. passive voice  & Num-passive voice verb(D) / W(D). A word is a passive voice verb iff word.deprel = `aux:pass'. We used Stanza for POS tagging. 
& 29. \textbf{SMOG readability index*} (smog) & Another index trying to estimate the years of education required to understand the text. We used TextSTAT to calculate this. \\
 \hline
 15. objectiﬁcation {[}skipped{]}   & {\it Missing computational description} \cite{zhou}. It is defined as an expression given (as an abstract notion, feeling, or ideal) in a form that can be experienced by others and externalizes one’s attitude. 
 & 30. automatic readability index* (ari) & Similar to Features (28) and (29). We also used TextSTAT for this. \\ 
\hline
\end{tabular}
\end{sidewaystable}


\begin{table}\centering

\caption{Behavior of features that show statistically significant differences between the truthful and deceptive classes. \textit{\# sig} is the number of datasets where the feature shows a significant difference. A positive number means the feature value is higher in the deceptive class. Emboldened features are those still qualified after the Bonferroni-Holm $p$-value adjustment.}
\label{tab:features_short}
\begin{tabular}{|r|l|c||r|l|c|} \hline
  & Feature            & \# sig &    & Feature       & \# sig \\\hline
1 & \textbf{word}      & $-3$     & 8  & modal verb     & $+3$     \\\hline
2 & \textbf{verb}      & $-3$     & 9  & certainty     & $+3$     \\\hline
3 & \textbf{sentence}  & $-3$     & 10 & \textbf{self ref}     & $-3$     \\\hline
4 & \textbf{sen\_len} & $-3$     & 11 & redundancy    & $-3$     \\\hline
5 & word\_len  & $-3$     & 12 & \textbf{SMOG}    & $+4$     \\\hline 
6 & \textbf{pausality}         & $-4$     & 13 & ARI & $-3$     \\\hline
7 & \textbf{modifier}          & $-3$     &    &        &     \\\hline
\end{tabular}
\end{table}

\subsection{Results: Selected Features}

We find that the following features are statistically lower for deceptive samples in three or more datasets:\footnote{The precise number of datasets with fewer is in Table~\ref{tab:features_short}.}
number of words,
number of verbs,
number of sentences,
sentence length,
word length, 
pausality (number of punctuation marks per sentence),
number of modifiers (adjectives and adverbs), 
self reference (number of first-person plural pronouns), 
redundancy (number of function words per sentence), and the Automatic Readability Index.  

We find that the following features are statistically higher for deceptive samples in three or more datasets: 
modal verbs,
certainty (number of words that have the certainty tag in the LIWC 2015 Dictionary per word),
and the SMOG readability index. We used Stanza~\cite{Stanza} for POS tagging. Removing positive affect, since it is covered by our LIWC list of features below, gives us 13 features. 

\section{Function Word \lowercase{{\large $n$}}-grams}
\label{sec:function_words}
In this section, we describe our initial analysis of function word $n$-grams for the same four public datasets as in the previous section of the appendix.


\subsection{Method}


\begin{figure}
\centering
\includegraphics[width=0.6\linewidth]{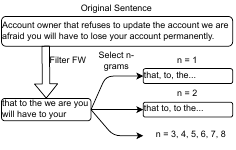}
\caption{An example of extracting function-word (FW) $n$-grams from a sentence.}\label{fig:FW_Ngram_exp}
\end{figure}

We combine function words and $n$-grams by looking for $n$-grams of function words that appear significantly more or less often in deceptive texts than truthful texts.

We first extract the function words from our texts using a list from the publicly-available PhishBench 2.0  \cite{zeng} and compute function-word $n$-grams for $n$ from one through eight (an example of this process is in Figure \ref{fig:FW_Ngram_exp}).\footnote{We also explored POS tagging, which produced similar results at greater computational cost.}

We then calculate the frequency of occurrence of every $n$-gram $x$ in each text $t$ using the formula
$$\textit{Occ$_n$}(x, t) = \sum_{s \in t} \frac{\#_x(s)}{|s|-n}$$
where  $|s|$ denotes the number of words in sentence $s$
of $t$, and $\#_x(s)$ denotes the number of times $x$ occurs in $s$.

We then ranked the function word $n$-grams by the difference of the aggregate occurrences in the two classes (truthful versus deceptive) and selected the top 100 for statistical significance testing. 
We then ran two-sample $t$-tests on each dataset, comparing occurrence scores between legitimate and deceptive texts, and identified ones that showed a consistent significant difference, both with and without Bonferroni-Holm correction.

\subsection{Results}

The results are in Table~\ref{tab:ngrams_short}.
Only two $n$-grams, ``do" and ``but" show significant differences between truthful/deceptive groups in all four datasets. Both unigrams ($n=1$) are more frequent in deceptive texts. Twenty $n$-grams, including ``I", ``they," ``is a," and ``at the," show significant differences between truthful and deceptive texts in three out of four datasets. After Bonferroni-Holm correction, 11 $n$-grams still qualified. 

\begin{table}\centering
\caption{Table of $n$-grams that show a consistent significant difference in \textit{Occ} in \textit{\# sig} datasets. A positive \textit{\# sig} means the $n$-grams is generally more frequent in deceptive texts. Emboldened unigrams are still qualified after the Bonferroni-Holm correction.}
\label{tab:ngrams_short}
\begin{tabular}{|l|l|c||l|l|c|} \hline
   & $n$-gram        & \# sig &    & $n$-gram          & \# sig \\\hline
1  & \textbf{I}    & $+3$          & 11 & their  & $+3$          \\\hline
2  & \textbf{they} & $+3$          & 12 & \textbf{if}     & $+3$          \\\hline
3  & out           & $+3$          & 13 & both   & $+3$          \\\hline
4  & is a          & $-3$          & 14 & at the & $+3$          \\\hline
5  & \textbf{only} & $+3$          & 15 & \textbf{will}   & $-3$          \\\hline
6  & do            & $+4$          & 16 & \textbf{me}     & $+3$          \\\hline
7  & \textbf{at}   & $+3$          & 17 & \textbf{but}    & $+4$          \\\hline
8  & about         & $+3$          & 18 & than            & $+3$          \\\hline
9  & \textbf{that} & $+3$          & 19 & \textbf{and}    & $-3$          \\\hline
10 & \textbf{them} & $+3$          & 20 & for             & $-3$ \\  \hline      
\end{tabular}
\end{table}

An individual dataset has thousands of FW $n$-grams, of which between 100 to 400 of them are significant discriminators of deceptive texts. In our experiments, 20 terms, 18 of them unigrams, show a common behavior across at least three out of the four datasets, and four across all four. After Bonferroni-Holm correction, 11 out of 20  are still qualified. We also notice that the lower the $n$, the higher the chance that the $n$-gram will show a significant difference between truthful and deceptive groups. 

\appendixsection{LIWC and BERTAA Features}\label{App:B}

The 55 stylistic features from~\cite{fabien2020bertaa} are listed below:
\begin{itemize}
    \item Length of text:   len-text
    \item Number of words:  len-words
    \item Average length of words: avg-len
    \item Number of short words:  num-short-w
    \item Proportion of digits and capital letters: per-digit, per-cap
    \item Individual letters and digits frequencies: f-a, f-b, f-c, f-d, f-e, f-f, f-g, f-h, f-i, f-j, f-k, f-l, f-m, f-n, f-o, f-p, f-q, f-r,f-s, f-t, f-u, f-v, f-w, f-x, f-y, f-z, f-0, f-1, f-2, f-3, f-4, f-5, f-6, f-7, f-8, f-9
    \item Hapax-legomena:  richness
    \item Frequency of 12 punctuation marks: f-e-0, f-e-1, f-e-2, f-e-3, f-e-4, f-e-5, f-e-6, f-e-7,
f-e-8, f-e-9, f-e-10, f-e-11
\end{itemize}

LIWC denotes the Linguistic Inquiry and Word Count program. It uses dictionaries of words that fit into different categories like tone\_pos (positive tone) to identify words in the text that fall into that category (happy, elated, excited, \ldots) and reports {\em percentage of words in the text} that fall into those categories except for WC (word count) and WPS (words per sentence). More information about LIWC can be found at \url{https://www.liwc.app/help}.

The 86 LIWC features that we used along with their abbreviations are as follows:
\begin{itemize}
    \item Summary Variables (8 features): Word count -- WC, Analytical thinking -- analytic, Clout -- clout, Authentic -- authentic, Emotional tone -- tone, Words per sentence -- WPS, Big words -- BigWords, Dictionary words -- Dic
    \item Linguistic Dimension (17 features): Total function words -- function, Total pronouns -- pronoun, Personal pronouns -- ppron, 1st person singular -- I, 1st person plural -- we, 3rd person plural -- they, Impersonal pronouns -- ipron, Determiners -- det, Articles -- article, Numbers -- number, Prepositions -- prep, Auxiliary verbs -- auxver, Adverbs --  adverb, Conjunctions -- conj, Negations -- negate, Common verbs -- verb, Quantities -- quantity
    \item Psychological Processes (28 features): Drives -- drives, Affiliation -- affiliation, Power -- power, Cognition -- cognition, All-or-none -- allnone, Cognitive processes -- cogproc, Insight -- insight, Causation -- cause, Discrepancy -- discrep, Tentative -- tentat, Certitude -- certitude, Differentiation -- differ, Affect -- affect, Positive tone -- tone\_pos, Negative tone -- tone\_neg, Emotion -- emotion, Positive emotion -- emo\_pos, Negative emotion -- emo\_neg, Anxiety -- emo\_anx, Sadness -- emo\_sad, Social processes -- Social, Social behavior -- socbehav, Prosocial behavior -- prosocial, Politeness -- polite, Moralization -- moral, Communication -- comm, Social referents -- socrefs, Friends -- friend
    \item Punctuation (5 features): All Punctuation -- allpunc, Apostrophes -- apostro, Periods -- period, Commas -- comma, Other punctuation -- OtherP 
\item Expanded Dictionary (28 features): Culture -- culture, Politics -- politic, Ethnicity -- ethnicity, Technology -- tech, Lifestyle -- lifestyle, Home -- home, Work -- work, Religion -- relig, Physical -- physical, Health
-- health, Mental health -- mental, Need -- need, Lack -- lack, Fulfilled -- fulfill, Risk -- risk, Curiosity -- curiosity, Allure -- allure, Perception -- perception, Attention -- attention, Motion -- motion, Space -- space, Visual
-- visual, Auditory -- auditory, Feeling -- feeling, Time -- time, Past focus -- focuspast, Present focus -- focuspresent, Future focus -- focusfuture
\end{itemize}

Note that a few of the features between these three lists: BERTAA, the 13 selected features from Table~\ref{tab:features_short} and LIWC list are duplicates. Specifically, 
WD from LIWC, words in Table~\ref{tab:features_short}, and len-text from BERTAA are duplicative, 
and WPS from LIWC, average sentence length in Table~\ref{tab:features_short} form another duplicate group.  These are removed by the colinearity check mentioned in the feature analysis section, Section~\ref{sec-cues}. 


\end{document}